%% file: main.tex
\documentclass{article}
\pdfoutput=1
\usepackage{biblatex}
\usepackage[final]{style}

\usepackage[utf8]{inputenc} 
\usepackage[T1]{fontenc}    
\usepackage{hyperref}       
\usepackage{url}            
\usepackage{booktabs}       
\usepackage{amsfonts}       
\usepackage{nicefrac}       
\usepackage{microtype}      
\usepackage{graphicx}
\usepackage{dsfont}
\usepackage{hyperref}
\usepackage{setspace}
\usepackage[table,xcdraw]{xcolor}

\title{Wide Activation for Efficient and Accurate Image Super-Resolution}
\author{
  Jiahui Yu\\
  \texttt{jyu79@illinois.edu}
  \And
  Yuchen Fan\\
  \texttt{yuchenf4@illinois.edu}
  \And
  Jianchao Yang\\
  \texttt{yangjianchao@bytedance.com}
  \And
  Ning Xu\\
  \texttt{ning.xu@snap.com}
  \And
  Zhaowen Wang\\
  \texttt{zhawang@adobe.com}
  \And
  Xinchao Wang\\
  \texttt{xwang135@stevens.edu}
  \And
  Thomas Huang\\
  \texttt{t-huang1@illinois.edu}
}

\begin{document}
\maketitle
\begin{abstract}
In this report we demonstrate that with same parameters and computational budgets, models with wider features before ReLU activation have significantly better performance for single image super-resolution (SISR). The resulted SR residual network has a slim identity mapping pathway with wider (\(2\times\) to \(4\times\)) channels before activation in each residual block. To further widen activation (\(6\times\) to \(9\times\)) without computational overhead, we introduce linear low-rank convolution into SR networks and achieve even better accuracy-efficiency tradeoffs. In addition, compared with batch normalization or no normalization, we find training with weight normalization leads to better accuracy for deep super-resolution networks. Our proposed SR network \textit{WDSR} achieves better results on large-scale DIV2K image super-resolution benchmark in terms of PSNR with same or lower computational complexity. Based on WDSR, our method also won 1st places in NTIRE 2018 Challenge on Single Image Super-Resolution in all three realistic tracks. Experiments and ablation studies support the importance of wide activation for image super-resolution. Code is released at: \url{https://github.com/JiahuiYu/wdsr_ntire2018}.
\end{abstract}

\input{sec1_intro.tex}

\input{sec2_related_work.tex}

\input{sec3_methods.tex}

\input{sec4_experiments.tex}

\section{Conclusions}

In this report, we introduce two super-resolution networks \textit{WDSR-A} and \textit{WDSR-B} based on the central idea of wide activation. We demonstrate in our experiments that with same parameter and computation complexity, models with wider features before ReLU activation have better accuracy for single image super-resolution. We also find training with weight normalization leads to better accuracy for deep super-resolution networks comparing to batch normalization or no normalization. The proposed methods may help to other low-level image restoration tasks like denoising and dehazing.

\printbibliography

\end{document}

%% file: sec1_intro.tex
\section{Introduction}

Deep convolutional neural networks (CNNs) have been successfully applied to the task of single image super-resolution (SISR)~\cite{kim2016accurate, lim2017enhanced, liu2016robust, 2018arXiv180208797Z}. SISR aims at recovery of a high resolution (HR) image from its low resolution (LR) counterpart (typically a bicubic downsampled version of HR). It has many applications in security, surveillance, satellite, medical imaging~\cite{peled2001superresolution, thornton2006sub} and can serve as a built-in module for other image restoration or recognition tasks~\cite{fan2018wide, liu2017robust, wang2016studying, yu2018free, yu2018generative}.
 
Previous image super-resolution networks including SRCNN~\cite{dong2014learning}, FSRCNN~\cite{dong2016accelerating}, ESPCN~\cite{shi2016real} utilized relatively shallow convolutional neural networks (with its depth from 3 to 5). They are inferior in accuracy compared with later proposed deep SR networks (e.g.,\ VDSR~\cite{kim2016accurate}, SRResNet~\cite{ledig2016photo} and EDSR~\cite{lim2017enhanced}). The increasing of depth brings benefits to representation power~\cite{cohen2016expressive, eldan2016power, liang2016deep, scarselli1998universal} but meanwhile under-use the feature information from shallow layers (usually represent low-level features). To address this issue, methods including SRDenseNet~\cite{tong2017image}, RDN~\cite{2018arXiv180208797Z}, MemNet~\cite{tai2017memnet} introduce various skip connections and concatenation operations between shallow layers and deep layers, formalizing holistic structures for image super-resolution.

In this work we address this issue in a different perspective. Instead of adding various shortcut connections, we conjecture that the non-linear ReLUs impede information flow from shallow layers to deeper ones~\cite{sandler2018inverted}. Based on residual SR network, we demonstrate that without additional parameters and computation, simply expanding features before ReLU activation leads to significant improvements for single image super-resolution, beating SR networks with complicated skip connections and concatenations including SRDenseNet~\cite{tong2017image} and MemNet~\cite{tai2017memnet}. The intuition of our work is that expanding features before ReLU allows more information pass through while still keeps highly non-linearity of deep neural networks. Thus low-level SR features from shallow layers may be easier to propagate to the final layer for better dense pixel value predictions.

\begin{figure}[t]
\centering
\includegraphics[width=\textwidth]{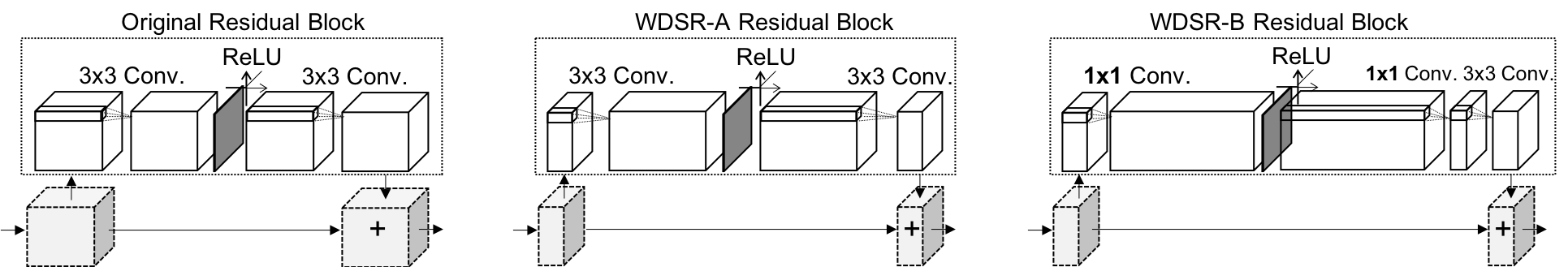}
\caption{\textbf{Left:} vanilla residual block. \textbf{Middle \textit{WDSR-A}:} residual block with wide activation. \textbf{Right \textit{WDSR-B}:} residual block with wider activation and linear low-rank convolution. We demonstrate different residual building blocks for image super-resolution networks. Compared with vanilla residual blocks used in EDSR~\cite{lim2017enhanced}, we introduce \textit{WDSR-A} which has a slim identity mapping pathway with wider (\(2\times\) to \(4\times\)) channels before activation in each residual block. We further introduce \textit{WDSR-B} with linear low-rank convolution stack and even widen activation (\(6\times\) to \(9\times\)) without computational overhead. In \textit{WDSR-A} and \textit{WDSR-B}, all ReLU activation layers are only applied between two wide features (features with larger channel numbers).}
\label{fig:wide}
\end{figure}

The central idea of wide activation leads us to explore efficient ways to expand features before ReLU, since simply adding more parameters is inefficient for real-time image SR scenarios~\cite{goto2014super}. We first introduce SR residual network \textit{WDSR-A}, which has a slim identity mapping pathway with wider (\(2\times\) to \(4\times\)) channels before activation in each residual block. However when the expansion ratio is above \(4\), channels of the identity mapping pathway have to be further slimmed and we find it dramatically deteriorates accuracy. Thus as the second step, we keep constant channel numbers of identity mapping pathway, and explore more efficient ways to expand features. We first consider group convolution~\cite{xie2017aggregated} and depthwise separable convolution~\cite{chollet2016xception}. However, we find both of them have unsatisfactory performance for the task of image super-resolution. To this end, we propose \textit{linear low-rank convolution} that factorizes a large convolution kernel into two low-rank convolution kernels. With wider activation and \textit{linear low-rank convolutions}, we construct our SR network \textit{WDSR-B}. It has even wider activation (\(6\times\) to \(9\times\)) without additional parameters or computation, and boosts accuracy further for image super-resolution. The illustration of \textit{WDSR-A} and \textit{WDSR-B} is shown in Figure~\ref{fig:wide}. Experiments show that wider activation consistently beats their baselines under different parameter budgets.

Additionally, compared with batch normalization~\cite{ioffe2015batch} or no normalization, we find training with weight normalization~\cite{salimans2016weight} leads to better accuracy for deep super-resolution networks. Previous works including EDSR~\cite{lim2017enhanced}, BTSRN~\cite{fan2017balanced} and RDN~\cite{2018arXiv180208797Z} found that batch normalization~\cite{ioffe2015batch} deteriorates the accuracy of image super-resolution, which is also confirmed in our experiments. We provide three intuitions and related experiments showing that batch normalization, due to 1) mini-batch dependency, 2) different formulations in training and inference and 3) strong regularization side-effects, is not suitable for training SR networks. However, with the increasing depth of neural networks for SR (e.g.\ MDSR~\cite{lim2017enhanced} has depth around 180), the networks without batch normalization become difficult to train. To this end, we introduce weight normalization for training deep SR networks. The weight normalization enables us to train SR network with an order of magnitude higher learning rate, leading to both faster convergence and better performance.

In summary, our contributions are as follows. 1) We demonstrate that in residual networks for SISR, wider activation has better performance with same parameter complexity. Without additional computational overhead, we propose network \textit{WDSR-A} which has wider (\(2\times\) to \(4\times\)) activation for better performance. 2) To further improve efficiency, we also propose \textit{linear low-rank convolution} as basic building block for construction of our SR network \textit{WDSR-B}. It enables even wider activation (\(6\times\) to \(9\times\)) without additional parameters or computation, and boosts accuracy further. 3) We suggest batch normalization~\cite{ioffe2015batch} is not suitable for training deep SR networks, and introduce weight normalization~\cite{salimans2016weight} for faster convergence and better accuracy. 4) We train proposed \textit{WDSR-A} and \textit{WDSR-B} built on the principle of wide activation with weight normalization, and achieve better results on large-scale DIV2K image super-resolution benchmark. Our method also won 1st places in NTIRE 2018 Challenge on Single Image Super-Resolution in all three realistic tracks.

%% file: sec2_related_work.tex
\section{Related Work}

\subsection{Super-Resolution Networks}
Deep learning-based methods for single image super-resolution significantly outperform conventional ones~\cite{park2003super, yang2010image} in terms of peak signal-to-noise ratio (PSNR) and structural similarity (SSIM). SRCNN~\cite{dong2014learning} was the first work utilizing an end-to-end convolutional neural network as a mapping function from LR images to their HR counterparts. Since then, various convolutional neural network architectures were proposed for improving the accuracy and efficiency. In this section, we review these approaches under several groups.

\textbf{Upsampling layers} Super-resolution involves upsampling operation of image resolution. The first super-resolution network SRCNN~\cite{dong2014learning} applied convolution layers on the pre-upscaled LR image. It is inefficient because all convolutional layers have to compute on high-resolution feature space, yielding \(S^2\) times computation than on low-resolution space, where \(S\) is the upscaling factor. To accelerate processing speed without loss of accuracy, FSRCNN~\cite{dong2016accelerating} utilized parametric deconvolution layer at the end of SR network~\cite{dong2016accelerating}, making all convolution layers compute on LR feature space. Another non-parametric efficient alternative is pixel shuffling~\cite{shi2016real} (a.k.a., sub-pixel convolution). Pixel shuffling is also believed to introduce less checkerboard artifacts~\cite{odena2016deconvolution} than the deconvolutional layer.

\textbf{Very deep and recursive neural networks} The depth of neural networks is of central importance for deep learning~\cite{he2016deep, simonyan2014very, szegedy2017inception}. It is also experimentally proved in single image super-resolution task~\cite{fan2017balanced, kim2016accurate, ledig2016photo, lim2017enhanced, tai2017memnet, tong2017image, 2018arXiv180208797Z}. These very deep networks (usually more than 10 layers) stack many small-kernel (i.e., \(3 \times 3\)) convolutions and have higher accuracy than shallow ones~\cite{dong2016accelerating, shi2016real}. However, the increasing depth of convolutional neural networks introduces over-parameterization and difficulty of training. To address these issues, recursive neural networks~\cite{kim2016deeply, tai2017image} are proposed by re-using weights repeatedly.

\textbf{Skip connections} On one hand, deeper neural networks have better performance in various tasks~\cite{simonyan2014very}, on the other hand low-level features are also important for image super-resolution task~\cite{2018arXiv180208797Z}. To address this contradictory,  VDSR~\cite{kim2016accurate} proposed a very deep VGG-like~\cite{simonyan2014very} network with global residual connection (i.e.\ identity skip connection) for SISR. SRResNet~\cite{ledig2016photo} proposed a ResNet-like~\cite{he2016deep} network. Densely connected networks~\cite{huang2017densely} are also adapted for SISR in SRDenseNet~\cite{tong2017image}. MemNet~\cite{tai2017memnet} integrated skip connections and recursive unit for low-level image restoration tasks. To further exploit the hierarchical features from all the convolutional layers, residual dense networks (RDN)~\cite{2018arXiv180208797Z} are proposed. All these works benefit from additional skip connections between different levels of features in deep neural networks.

\textbf{Normalization layers} As image super-resolution networks going deeper and deeper (from 3-layer SRCNN~\cite{dong2014learning} to 160-layer MDSR~\cite{lim2017enhanced}), training becomes more difficult. Batch normalization layers are one of the cures for this problem in many tasks~\cite{he2016deep, szegedy2017inception}. It is also introduced in SISR networks in SRResNet~\cite{ledig2016photo}. However, empirically it is found that batch normalization~\cite{ioffe2015batch} hinders the accuracy of image super-resolution. Thus, in recent image SR networks~\cite{fan2017balanced, lim2017enhanced, 2018arXiv180208797Z}, batch normalization is abandoned.

\subsection{Parameter-Efficient Convolutions}

In this subsection, we also review several related methods proposed for improving efficiency of convolutions.

\textbf{Flattened convolution} Flattened convolutions~\cite{jin2014flattened} consist of consecutive sequence of one-dimensional filters across all directions in 3D space (lateral, vertical and horizontal) to approximate conventional convolutions. The number of parameters in flattened convolution decreases from \(XYC\) to \(X+Y+C\), where \(C\) is the number of input planes, \(X\) and \(Y\) denote filter width and height.

\textbf{Group convolution} Group convolutions~\cite{xie2017aggregated} divide features into groups channel-wisely and perform convolutions inside the group individually, followed by a concatenation to form the final output. In group convolutions, the number of parameters can be reduced by \(g\) times, where \(g\) is the group number. Group convolutions are the key components to many efficient models (e.g.\ ResNeXt~\cite{xie2017aggregated}).

\textbf{Depthwise separable convolution} Depthwise separable convolution is a stack of depthwise convolution (i.e.\ a spatial convolution performed independently over each channel of an input) followed by a pointwise convolution (i.e.\ a 1x1 convolution) without non-linearities. It can also be viewed as a specific type of group convolution where the number of groups \(g\) is the number of channels. The depthwise separable convolution formulates the basic architecture in many efficient models including Xception~\cite{chollet2016xception}, MobileNet~\cite{howard2017mobilenets} and MobileNetV2~\cite{2018arXiv180104381S}.

\textbf{Inverted residuals} Another work~\cite{2018arXiv180104381S} expands features before activation for image recognition tasks (named inverted residuals). The intermediate expansion layer uses lightweight depthwise convolutions to filter features as a source of non-linearity. The inverted residual shares similar merits with our proposed wide activation, however we found the inverted residual proposed in~\cite{2018arXiv180104381S} has unsatisfactory performance on the task of image SR. In this work we mainly explore different network architectures to improve the accuracy and efficiency for the task of image super-resolution with the central idea of wide activation.

%% file: sec3_methods.tex
\section{Proposed Methods}
\subsection{Wide Activation: \textit{WDSR-A}}
In this part, we mainly describe how we expand features before ReLU activation layer without computational overhead. We consider the effects of wide activation inside a residual block. A naive way is to directly add channel numbers of all features. However, it proves nothing except that more parameters lead to better performance. Thus, in this section, we design our SR network to study the importance of wide features before activation with \textit{same parameter and computational budgets}. Our first step towards wide activation is extremely simple: we slim the features of residual identity mapping pathway while expand the features before activation, as shown in Figure~\ref{fig:wide}.

Two-layer residual blocks are specifically studied following baseline EDSR~\cite{lim2017enhanced}. Assume the width of identity mapping pathway (Fig.~\ref{fig:network}) is \(w_1\) and width before activation inside residual block is \(w_2\). We introduce expansion factor before activation as \(r\) thus \(w_2 = r \times w_1\). In the vanilla residual networks (e.g.,\ used in EDSR and MDSR) we have \(w_2 = w_1\) and the number of parameters are \(2 \times w_1^2 \times k^2\) in each residual block. The computational (Mult-Add operations) complexity is a constant scaling of parameter numbers when we fix the input patch size. To have same complexity \(w_1^2 = \hat{w_1} \times \hat{w_2} = r \times \hat{w_1}^2\), the residual identity mapping pathway need to be slimmed as a factor of \(\sqrt{r}\) and the activation can be expanded with \(\sqrt{r}\) times meanwhile.

This simple idea forms our first widely-activated SR network \textit{WDSR-A}. Experiments show that \textit{WDSR-A} is extremely effective for improving accuracy of SISR when \(r\) is between 2 to 4. However, for \(r\) larger than this threshold the performance drops quickly. This is likely due to the identity mapping pathway becoming too slim. For example, in our baseline EDSR (16 residual blocks with 64 filters) for \(\times 3\) super-resolution, when \(r\) is beyond 6, \(w_1\) will be even smaller than the final HR image representation space \(S^2*3\) (we use pixel shuffle as upsampling layer) where \(S\) is the scaling factor and 3 represents RGB. Thus we seek for parameter-efficient convolution to further improve accuracy and efficiency with wider activation.

\subsection{Efficient Wider Activation: \textit{WDSR-B}}
To address the above limitation, we keep constant channel numbers of identity mapping pathway, and explore more efficient ways to expand features. Specifically we consider \(1 \times 1\) convolutions. \(1 \times 1\) convolutions are widely used for channel number expansion or reduction in ResNets~\cite{he2016deep}, ResNeXts~\cite{xie2017aggregated} and MobileNetV2~\cite{2018arXiv180104381S}. In \textit{WDSR-B} (Fig. \ref{fig:wide}) we first expand channel numbers by using \(1 \times 1\) and then apply non-linearity (ReLUs) after the convolution layer. We further propose an efficient \textit{linear low-rank convolution} which factorizes a large convolution kernel to two low-rank convolution kernels. It is a stack of one \(1 \times 1\) convolution to reduce number of channels and one \(3 \times 3\) convolution to perform spatial-wise feature extraction. We find adding ReLU activation in \textit{linear low-rank convolutions} significantly reduces accuracy, which also supports wide activation hypothesis.

\subsection{Weight Normalization vs. Batch Normalization}

In this part, we mainly analyze the different purposes and effects of batch normalization (BN)~\cite{ioffe2015batch} and weight normalization (WN)~\cite{salimans2016weight}. We offer three intuitions why batch normalization is not appropriate for image SR tasks. Then we demonstrate that weight normalization does not have these drawbacks like BN, and it can be effectively used to ease the training difficulty of deep SR networks.

\textbf{Batch normalization} BN re-calibrates the mean and variance of intermediate features to solve the problem of \textit{internal covariate shift}~\cite{ioffe2015batch} in training deep neural networks. It has different formulations in training and testing. For simplicity, here we ignore the re-scaling and re-centering learnable parameters of BN. During training, features in each layer are normalized with mean and variance of the current training mini-batch:
\begin{equation}
\hat x_B = \frac{x_B - E_B[x_B]}{\sqrt{Var_B[x_B]+\epsilon}},
\end{equation}
where \(x_B\) is the features of current training batch, \(\epsilon\) is a small value (e.g.\ 1e-5) to avoid zero-division. The first order and second order statistics are then updated to global statistics in a moving average way:

\begin{equation}
E[x] \leftarrow E_B[x_B]\\,
\end{equation}
\begin{equation}
Var[x] \leftarrow Var_B[x_B], 
\end{equation}
where \(\leftarrow\) means assigning moving average. During inference, these global statistics are used instead to normalize the features:
\begin{equation}
\hat x_{test} = \frac{x_{test} - E[x]}{\sqrt{Var[x]+\epsilon}}.
\end{equation}

As shown in the formulations of BN, it will cause following problems.
1) For image super-resolution, commonly only small image patches (e.g.\ \(48 \times 48\)) and small mini-batch size (e.g.\ 16) are used to speedup training~\cite{fan2017balanced, kim2016accurate, ledig2016photo, lim2017enhanced, tai2017memnet, tong2017image, 2018arXiv180208797Z}, thus the mean and variance of small image patches differ a lot among mini-batches, making theses statistics unstable, which is demonstrated in the section of experiments.
2) BN is also believed to act as a regularizer and in some cases can eliminate the need for Dropout~\cite{ioffe2015batch}. However, it is rarely observed that SR networks overfit on training datasets. Instead, many kinds of regularizers, for examples, weight decaying and dropout, are not adopted in SR networks~\cite{fan2017balanced, kim2016accurate, ledig2016photo, lim2017enhanced, tai2017memnet, tong2017image, 2018arXiv180208797Z}.
3) Unlike image classification tasks where softmax (scale-invariant) is used at the end of networks to make prediction, for image SR, the different formulations of training and testing may deteriorate the accuracy for dense pixel value predictions.

\textbf{Weight normalization} Weight normalization, on the other hand, is a reparameterization of the weight vectors in a neural network that decouples the length of those weight vectors from their direction. It does not introduce dependencies between the examples in a mini-batch, and has the same formulation in training and testing. Assume the output \(\mathbf{y}\) is with the form:
\begin{equation}
\mathbf{y} = \mathbf{w} \cdot \mathbf{x} + b, 
\end{equation}
where \(\mathbf{w}\) is a k-dimensional weight vector, \(b\) is a scalar bias term, \(\mathbf{x}\) is a k-dimensional vector of input features. WN re-parameterizes the weight vectors in terms of the new parameters using
\begin{equation}
\mathbf{w} = \frac{g}{||\mathbf{v}||} \mathbf{v}, 
\end{equation}
where v is a k-dimensional vector, g is a scalar, and \(||\mathbf{v}||\) denotes the Euclidean norm of \(\mathbf{v}\). With this formalization, we will have \(||\mathbf{w}|| = g\), independent of parameters \(\mathbf{v}\). As shown in~\cite{salimans2016weight}, the decouples of length and direction speed up convergence of deep neural networks. And more importantly, for image SR, it does not introduce troubles of BN as described above, since it is just a reparameterization technique and has exact same representation ability.

It is also noteworthy that introducing WN allows training with higher learning rate (i.e.\ \(10 \times\)), and improves both training and testing accuracy.

\subsection{Network Structure}

\begin{figure}[h]
\centering
\includegraphics[width=\textwidth]{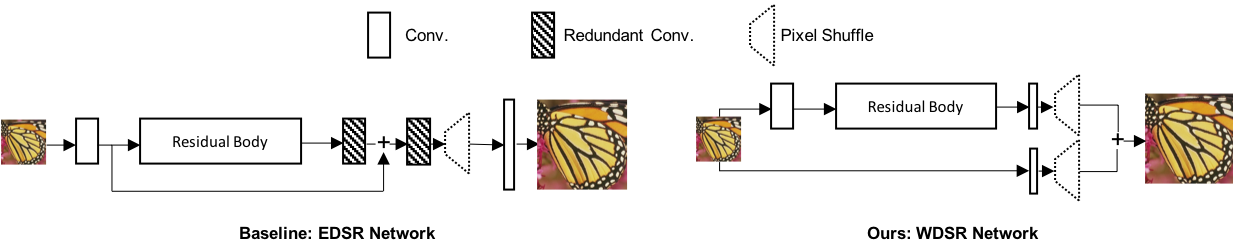}
\caption{Demonstration of our simplified SR network compared with EDSR~\cite{lim2017enhanced}.}
\label{fig:network}
\end{figure}

In this part, we overview the \textit{WDSR} network architectures. We made two major modifications based on EDSR~\cite{lim2017enhanced} super-resolution network.

\textbf{Global residual pathway} Firstly we find that the global residual pathway is a linear stack of several convolution layers, which is computational expensive. We argue that these linear convolutions are redundant (Fig.~\ref{fig:network}) and can be absorbed into residual body to some extent. Thus, we slightly modify the network structure and use single convolution layer with kernel size \(5 \times 5\) that directly take \(3 \times H \times W\) LR RGB image/patch as input and output \(3S^2 \times H \times W\) HR counterparts, where \(S\) is the scale. This results in less parameters and computation. In our experiments we have not found any accuracy drop with our simpler form.

\textbf{Upsampling layer} Different from previous state-of-the-arts~\cite{lim2017enhanced, 2018arXiv180208797Z} where one or more convolutional layers are inserted after upsampling, our proposed \textit{WDSR} extracts all features in low-resolution stage (Fig.~\ref{fig:network}). Empirically we find it does not affect accuracy of SR networks while improves speed by a large margin.

%% file: sec4_experiments.tex
\section{Experimental Results}

We train our models on DIV2K dataset~\cite{timofte2017ntire} since the dataset is relatively large and contains high-quality (2K resolution) images. The default splits of DIV2K dataset consist 800 training images, 100 validation images and 100 testing images. We use 800 training images for training and 10 validation images for validation during training. The trained models are evaluated on 100 validation images (testing images are not publicly available) of DIV2K dataset. We mainly measure PSNR on RGB space. ADAM optimizer~\cite{kingma2014adam} is used with \(\beta_1 = 0.9\), \(\beta_2 = 0.999\) and \(\epsilon = 10^{-8}\). The batch size is set to 16. The learning rate is initialized the maximum convergent value (10-4 for models without weight normalization and 10-3 for models with weight normalization). The learning rate is halved at every \(2 \times 10^5\) iterations.

We crop \(96 \times 96\) RGB input patches from HR image and its bicubic downsampled image as training output-input pairs. Training data is augmented with random horizontal flips and rotations following common data augmentation methods~\cite{fan2017balanced, lim2017enhanced}. During training, the input images are also subtracted with the mean RGB values of the DIV2K training images.

\subsection{Wide and Efficient Wider Activation:}

In this part, we show results of baseline model EDSR~\cite{lim2017enhanced} and our proposed \textit{WDSR-A} and \textit{WDSR-B} for the task of image bicubic x2 super-resolution on DIV2K dataset. To ensure fairness, each model is evaluated at different parameters and computational budgets by controlling the number of residual blocks with fixed number of channels. The results are shown in Table~\ref{figs:wide_activation}. We compare each model with its number of residual blocks. The results suggest that our proposed \textit{WDSR-A} and \textit{WDSR-B} have better accuracy and efficiency than EDSR~\cite{lim2017enhanced}. \textit{WDSR-B} with wider activation also has better or similar performance compared with \textit{WDSR-A}, which supports our wide activation hypothesis and demonstrates the effectiveness of our proposed \textit{linear low-rank convolution}.

\input{tabs/sr_edsr_wdsr.tex}

\subsection{Normalization layers:}

\begin{figure}[h]
\centering
\includegraphics[width=0.48\textwidth]{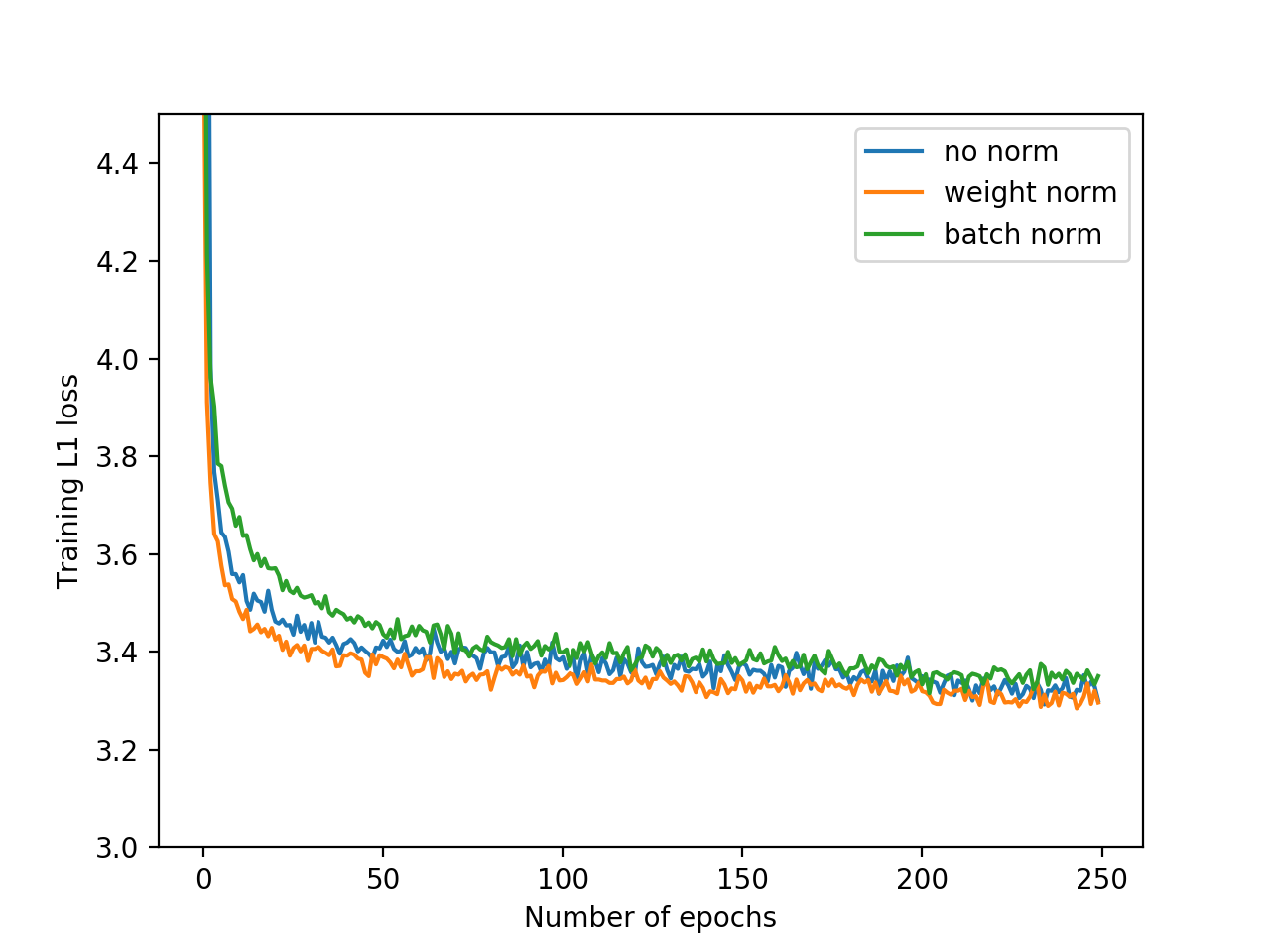}
\includegraphics[width=0.48\textwidth]{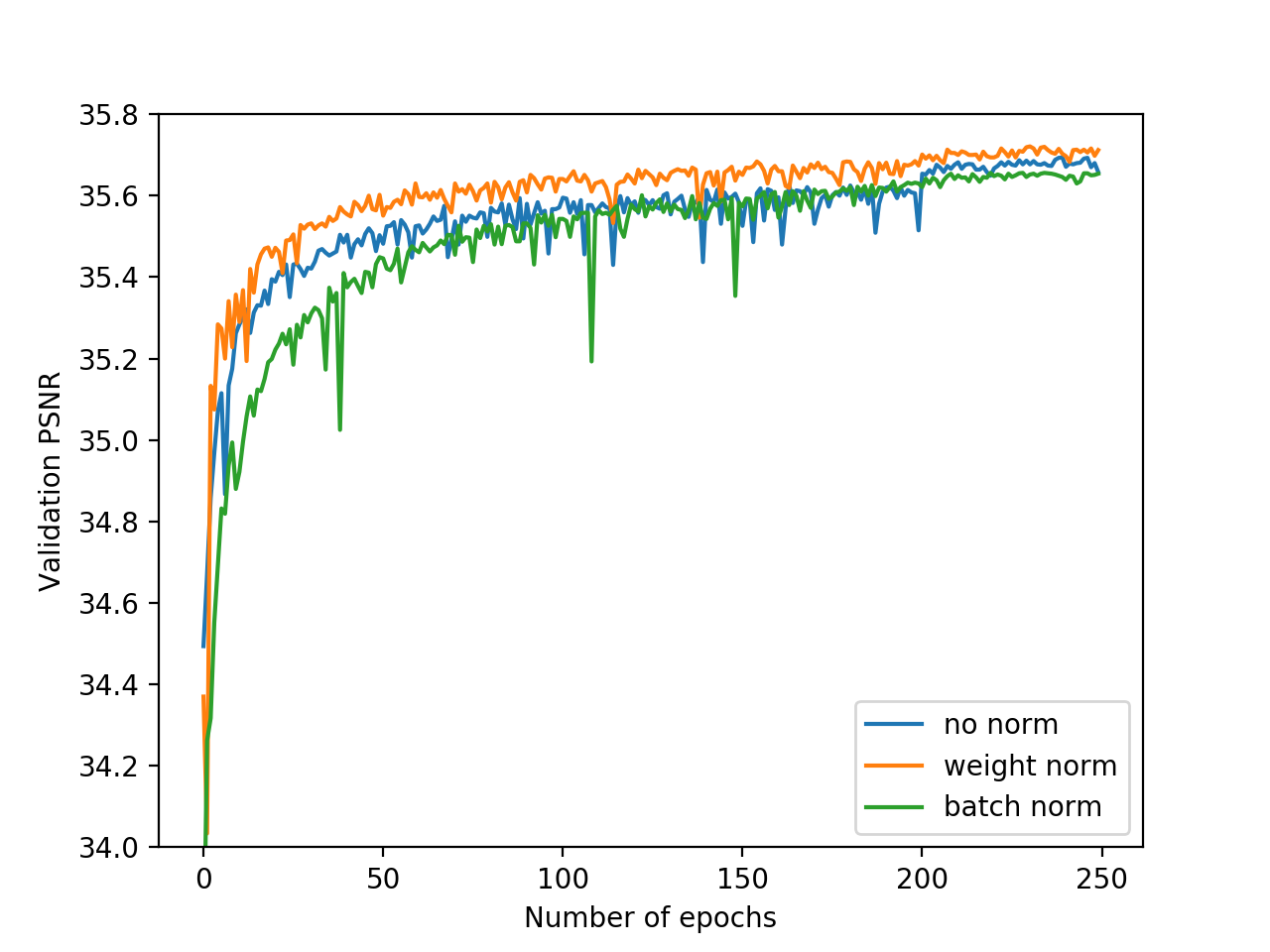}
\caption{Training L1 loss and validation PSNR of same model trained with weight normalization, batch normalization or no normalization.}
\label{figs:weight_norm}
\end{figure}

We also demonstrate the effectiveness of weight normalization for improved training of SR networks. We compare the training and testing accuracy (PSNR) when train the same model with different normalization methods, i.e. weight normalization, batch normalization or no normalization. The results in Figure~\ref{figs:weight_norm} show that the model trained with weight normalization has faster convergence and better accuracy. The model trained with batch normalization is unstable during testing, which is likely due to different formulations of BN in training and testing.

\begin{figure}[h]
\centering
\includegraphics[width=0.48\textwidth]{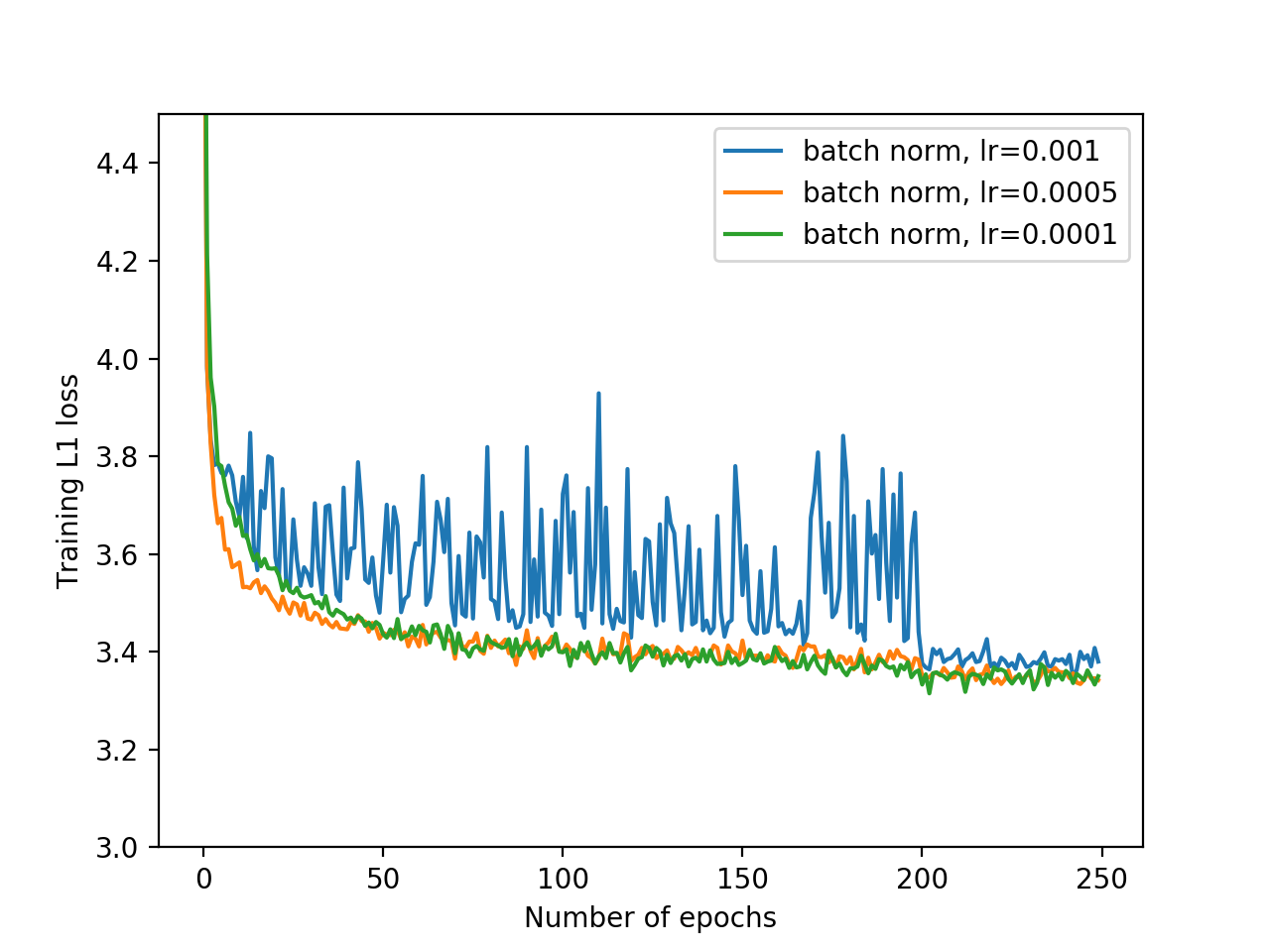}
\includegraphics[width=0.48\textwidth]{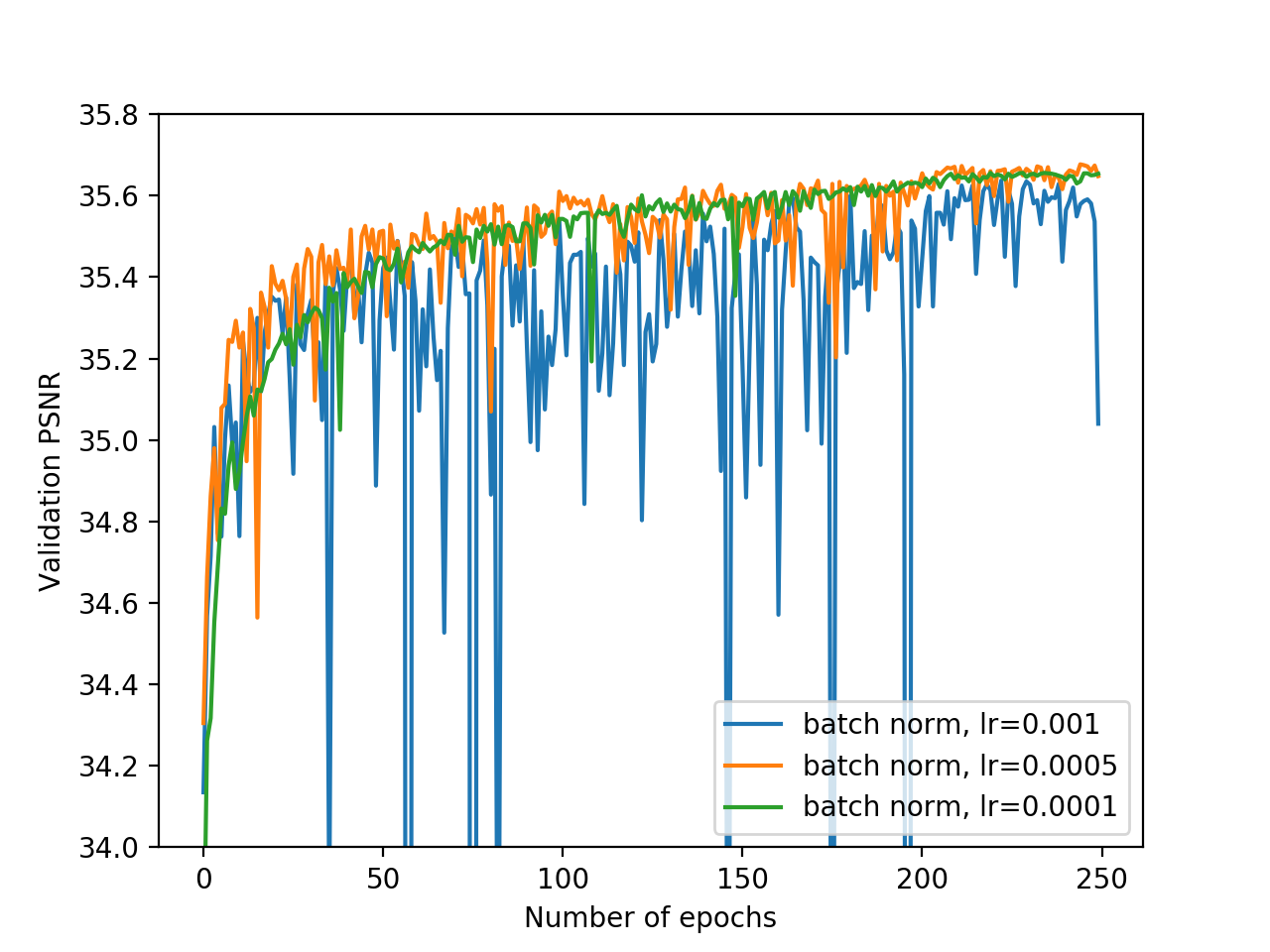}
\caption{Training L1 loss and validation PSNR of model trained with batch normalization but different learning rates.}
\label{figs:batch_norm}
\end{figure}

To further study whether this is because the learning rate is too large for models trained with batch normalization, we also train the same model with different learning rates. The results are shown in Figure~\ref{figs:batch_norm}. Even with \(lr = 10^{-4}\) when the training curves are stable, the validation PSNR is still not stable across training. 

%% file: tabs/sr_edsr_wdsr.tex

\begin{table}[]
\centering
\begin{tabular}{ccccccc}
\hline
\multicolumn{1}{|c|}{Residual Blocks}  & \multicolumn{3}{c|}{1}                                                                                                                              & \multicolumn{3}{c|}{3}                                                                                                                              \\ \hline
\multicolumn{1}{|c|}{Networks}         & \multicolumn{1}{c|}{EDSR}   & \multicolumn{1}{c|}{WDSR-A}                               & \multicolumn{1}{c|}{WDSR-B}                               & \multicolumn{1}{c|}{EDSR}   & \multicolumn{1}{c|}{WDSR-A}                               & \multicolumn{1}{c|}{WDSR-B}                               \\ \hline
\multicolumn{1}{|c|}{Parameters}       & \multicolumn{1}{c|}{0.26M}   & \multicolumn{1}{c|}{{\color[HTML]{333333} \textbf{0.08M}}} & \multicolumn{1}{c|}{{\color[HTML]{333333} \textbf{0.08M}}} & \multicolumn{1}{c|}{0.41M}   & \multicolumn{1}{c|}{{\color[HTML]{333333} \textbf{0.23M}}} & \multicolumn{1}{c|}{{\color[HTML]{333333} \textbf{0.23M}}} \\ \hline
\multicolumn{1}{|c|}{DIV2K (val) PSNR} & \multicolumn{1}{c|}{33.210} & \multicolumn{1}{c|}{{\color[HTML]{0000FF} 33.323}}        & \multicolumn{1}{c|}{{\color[HTML]{FF0000} 33.434}}        & \multicolumn{1}{c|}{34.043} & \multicolumn{1}{c|}{{\color[HTML]{0000FF} 34.163}}        & \multicolumn{1}{c|}{{\color[HTML]{FF0000} 34.205}}        \\ \hline
                                       &                             &                                                           &                                                           &                             &                                                           &                                                           \\ \hline
\multicolumn{1}{|c|}{Residual Blocks}  & \multicolumn{3}{c|}{5}                                                                                                                              & \multicolumn{3}{c|}{8}                                                                                                                              \\ \hline
\multicolumn{1}{|c|}{Networks}         & \multicolumn{1}{c|}{EDSR}   & \multicolumn{1}{c|}{WDSR-A}                               & \multicolumn{1}{c|}{WDSR-B}                               & \multicolumn{1}{c|}{EDSR}   & \multicolumn{1}{c|}{WDSR-A}                               & \multicolumn{1}{c|}{WDSR-B}                               \\ \hline
\multicolumn{1}{|c|}{Parameters}       & \multicolumn{1}{c|}{0.56M}   & \multicolumn{1}{c|}{{\color[HTML]{333333} \textbf{0.37M}}} & \multicolumn{1}{c|}{{\color[HTML]{333333} \textbf{0.37M}}} & \multicolumn{1}{c|}{0.78M}   & \multicolumn{1}{c|}{{\color[HTML]{333333} \textbf{0.60M}}} & \multicolumn{1}{c|}{{\color[HTML]{333333} \textbf{0.60M}}} \\ \hline
\multicolumn{1}{|c|}{DIV2K (val) PSNR} & \multicolumn{1}{c|}{34.284} & \multicolumn{1}{c|}{{\color[HTML]{0000FF} 34.388}}        & \multicolumn{1}{c|}{{\color[HTML]{FF0000} 34.409}}        & \multicolumn{1}{c|}{34.457} & \multicolumn{1}{c|}{{\color[HTML]{FF0000} 34.541}}        & \multicolumn{1}{c|}{{\color[HTML]{0000FF} 34.536}}        \\ \hline
\end{tabular}
\vspace{0.2cm}
\caption{Model comparisons at different parameters budgets by controlling the number of residual blocks with fixed number of channels. We mainly compare the number of parameters and validation PSNR to measure efficiency and accuracy.}
\label{figs:wide_activation}
\end{table}